\documentclass[11pt]{article}

\usepackage{dsfont}
\usepackage[dvipsnames]{xcolor}
\usepackage[preprint]{acl}
\usepackage{tablefootnote}
\usepackage{times}
\usepackage{latexsym}
\usepackage{booktabs}   
\usepackage{multirow}   
\usepackage{pifont}     
\usepackage[table]{xcolor} 
\usepackage{adjustbox}  
\usepackage{booktabs}
\usepackage{multirow}
\usepackage{adjustbox}
\usepackage[table]{xcolor}
\newcommand{\cmark}{\ding{51}}%
\newcommand{\xmark}{\ding{55}}%
\usepackage{graphicx} 
\usepackage{makecell}
\usepackage[T1]{fontenc}
\usepackage{verbatim}
\usepackage[table]{xcolor}
\usepackage{booktabs}
\usepackage[normalem]{ulem}
\usepackage[utf8]{inputenc}
\usepackage[dvipsnames]{xcolor} 
\usepackage{amssymb} 
\usepackage{colortbl} 
\usepackage[dvipsnames]{xcolor} 
\usepackage{amsmath}
\newcommand{\inc}[1]{%
    \textcolor{OliveGreen}{\textit{\footnotesize (#1)}}%
}
\usepackage{microtype}
\usepackage{mdframed}
\usepackage{inconsolata}
\usepackage{xcolor}
\definecolor{Emerald}{RGB}{0, 168, 107}

\usepackage{graphicx}
\usepackage[dvipsnames]{xcolor}

\title{Region-Grounded Report Generation for 3D Medical Imaging:\\A Fine-Grained Dataset and Graph-Enhanced Framework}

\author{
\\
  \textbf{Cong Huy Nguyen\textsuperscript{1,$*$}},
  \textbf{Son Dinh Nguyen\textsuperscript{1,$*$}},
  \textbf{Guanlin Li\textsuperscript{2}},
  \textbf{Tuan Dung Nguyen\textsuperscript{1}},
\\
  \textbf{Aditya Narayan Sankaran\textsuperscript{2}},
  \textbf{Mai Huy Thong\textsuperscript{3}},
  \textbf{Thanh Trung Nguyen\textsuperscript{3}},
  \textbf{Mai Hong Son\textsuperscript{3}},
\\
  \textbf{Reza Farahbakhsh\textsuperscript{2}},
  \textbf{Phi Le Nguyen\textsuperscript{1,$\dagger$}},
  \textbf{Noel Crespi\textsuperscript{2,$\dagger$}}
\\
\\
  \textsuperscript{1}AI4LIFE, Hanoi University of Science and Technology, Vietnam\\
  \small\texttt{\{huy.nc235504, son.dn225997, dung.nt232198m\}@sis.hust.edu.vn, lenp@soict.hust.edu.vn}
\\
  \textsuperscript{2}SAMOVAR, Télécom SudParis, Institut Polytechnique de Paris, France\\
  \small\texttt{\{guanlin\_li, aditya-narayan.sankaran, reza.farahbakhsh, noel.crespi\}@telecom-sudparis.eu} 
\\
  \textsuperscript{3}108 Military Central Hospital, Vietnam\\
  \small\texttt{maihuythong27121995@gmail.com, trungntc10@benhvien108.vn, alex.hong.son@gmail.com}
}

\renewcommand{\inc}[1]{\footnotesize{\textcolor{ForestGreen}{#1}}}

\begin{document}
\maketitle
\begingroup\def\thefootnote{$*$}\footnotetext{Equal contribution.}\endgroup
\begingroup\def\thefootnote{$\dagger$}\footnotetext{Corresponding authors.}\endgroup
\begin{abstract}
Automated medical report generation for 3D PET/CT imaging is fundamentally challenged by the high-dimensional nature of volumetric data and a critical scarcity of annotated datasets, particularly for low-resource languages. Current ``black-box'' methods map whole volumes to reports, ignoring the clinical workflow of analyzing localized Regions of Interest (RoIs) to derive diagnostic conclusions. In this paper, we bridge this gap by introducing \textit{VietPET-RoI}, the first large-scale 3D PET/CT dataset with fine-grained RoI annotation for a low-resource language, comprising 600 PET/CT samples and 1,960 manually annotated RoIs, paired with corresponding clinical reports. Furthermore, to demonstrate the utility of this dataset, we propose \textit{HiRRA}, a novel framework that mimics the professional radiologist diagnostic workflow by employing graph-based relational modules to capture dependencies between RoI attributes. This approach shifts from global pattern matching toward localized clinical findings. Additionally, we introduce new clinical evaluation metrics, namely \textit{RoI Coverage} and \textit{RoI Quality Index}, that measure both RoI localization accuracy and attribute description fidelity using LLM-based extraction. Extensive evaluation demonstrates that our framework achieves SOTA performance, surpassing existing models by 19.7\% in BLEU and 4.7\% in ROUGE-L, while achieving a remarkable 45.8\% improvement in clinical metrics, indicating enhanced clinical reliability and reduced hallucination. Our code and dataset are available on \href{https://github.com/Etdihatthoc/VietPET-RoI_ACL2026.git}{GitHub}.
\end{abstract}

\newpage
\section{Introduction}
\vspace{-0.1cm}

\begin{table*}[t]
\centering
\caption{\textbf{Comparison with existing PET/CT benchmarks.} VietPET-RoI uniquely provides both full reports and fine-grained RoI annotations (3D Boxes, Attributes).}

\label{tab:dataset_comparison}
\begin{adjustbox}{width=0.95\textwidth}
\begin{tabular}{l c c c c c c c c c c c}
\toprule
\multirow{2}{*}{\textbf{Dataset}} & 
\multirow{2}{*}{\textbf{Year}} &
\multirow{2}{*}{\textbf{Lang}} & 
\multicolumn{4}{c}{\textbf{Dataset Profile}} & 
\multicolumn{5}{c}{\textbf{Annotation Granularity}} \\ 
\cmidrule(lr){4-7} \cmidrule(lr){8-12} 
 & & & 
 \textbf{Size} & \textbf{Disease} & \textbf{Dim} & \textbf{Public} & 
 \textbf{Report} & \textbf{Metadata} & \textbf{RoI BBox} & \textbf{RoI Attrs} & \textbf{Phys/Path} \\ 
\midrule
ViMed-PET~\cite{nguyen2026toward} & 2025 & VN & 2,757 & Multi & 3D & \cmark & \cmark & \cmark & \xmark & \xmark & \xmark \\
PETRG-Lym~\cite{jiao2025vision}   & 2025 & CN & 824   & Single & 3D & \xmark & \cmark & \xmark & \xmark & \xmark & \xmark \\ 
PET2Rep~\cite{zhang2026pet2rep}   & 2025 & CN & 565   & Multi & 2D & \cmark & \cmark & \xmark & \xmark & \xmark & \xmark \\
AutoPET-RG~\cite{jiao2025vision}  & 2025 & CN & 135   & Single & 3D & \cmark & \cmark & \xmark & \xmark & \xmark & \xmark \\
\midrule
\rowcolor{gray!15} 
\textbf{VietPET-RoI (Ours)} & \textbf{2026} & \textbf{VN} & \textbf{600}\footnotemark & \textbf{Multi} & \textbf{3D} & \textbf{\cmark} & \textbf{\cmark} & \textbf{\cmark} & \textbf{\cmark} & \textbf{\cmark} & \textbf{\cmark} \\ 
\bottomrule
\end{tabular}
\end{adjustbox}
\begin{flushleft}
\footnotesize{
\textbf{Lang}: Language (VN: Vietnamese, CN: Chinese). 
\textbf{Metadata}: De-identified patient metadata.
\textbf{RoI BBox}: Manual 3D Bounding Box annotation. 
\textbf{RoI Attrs}: Rich structured attributes (Density, Size, etc). 
\textbf{Phys/Path}: Physiological/Pathological uptake.
}

\end{flushleft}
\vspace{-0.6cm}
\end{table*}

\begin{figure}[t]
    \centering
    \includegraphics[width=0.47\textwidth]{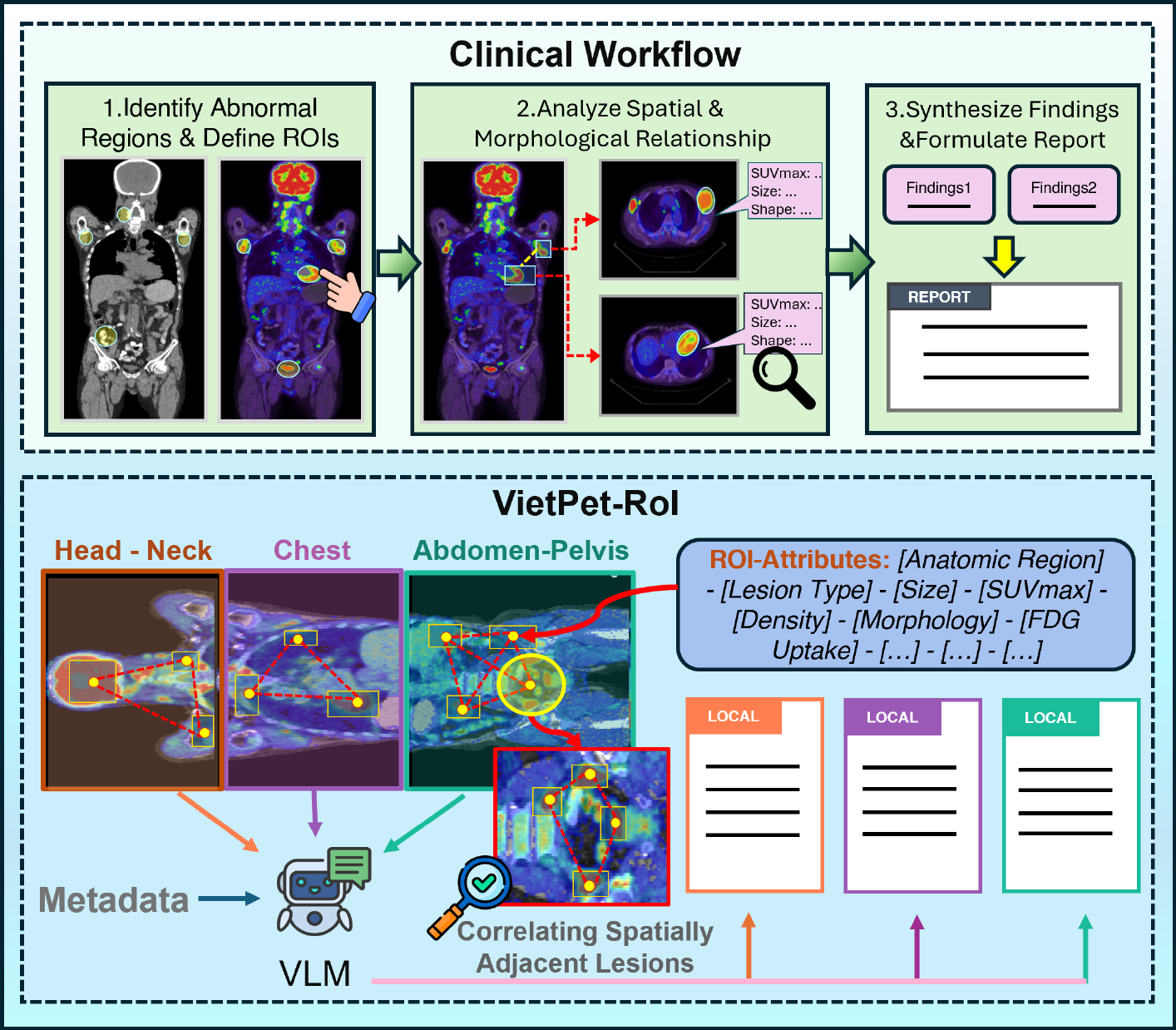}
    \vspace{-0.1cm}
    \caption{\textbf{Illustration of VietPET-RoI annotation.} Following doctors' conventional workflow, VietPET-RoI provides hierarchical annotations at both region-level and RoI-level with structured clinical attributes.}
    \label{fig:petct_example}

    \vspace{-0.3cm}
\end{figure}
Recent advances in Vision-Language Models (VLMs) have driven significant progress in healthcare AI, enabling the automated generation of clinical reports from medical images. Contemporary medical VLMs, such as LLaVA-Med \citep{li2023llava}, M3D-LaMed \citep{bai2024m3d}, and RadFM \citep{wu2025towards}, have shown impressive capabilities in interpreting diagnostic imaging. 

However, despite these general advancements, automated report generation for 3D PET/CT remains in its infancy. Current models exhibit suboptimal accuracy and are prone to significant hallucinations \cite{gu2026medvh}, with state-of-the-art VLMs falling considerably short of real-world clinical requirements \cite{zhang2026pet2rep}, highlighting substantial challenges in 3D multimodal analysis. 
We posit that this limitation stems from a fundamental methodological divergence. Existing PET/CT report generation models predominantly rely on an end-to-end paradigm, attempting to map complex, high-dimensional whole-volume scans directly to a final text report \cite{messina2022survey}. This ``black-box'' strategy ignores the intrinsic complexity of PET/CT data. In practice, radiologists do not interpret a 3D volume as a single monolithic input; instead, they systematically identify specific Regions of Interest (RoIs), evaluate their individual attributes, and analyze the spatial and physiological inter-relationships between these abnormalities \cite{waite2019analysis}. Only after this granular synthesis do they derive diagnostic conclusions and draft a formal report (shown in Figure~\ref{fig:petct_example}). Consequently, the current end-to-end training paradigm lacks the clinical inductive bias, leading to reports that lack both precision and interpretability.

To address this challenge, two fundamental components are required: (i) a dataset with fine-grained RoI-level annotations to provide a basis for grounded learning \cite{xie2025medtrinity, boecking2022making, de2025padchest}, and (ii) a model architecture that replicates the hierarchical reasoning process of a medical expert \cite{zhang2024hierarchical}. However, existing PET/CT datasets and models \cite{bai2024m3d, nguyen2026toward} fail to meet these requirements, particularly for low-resource languages like Vietnamese.
\footnotetext{600 region-level samples derived from 200 patients.}

In this paper, we bridge this research gap by introducing VietPET-RoI, the first 3D PET/CT dataset featuring fine-grained RoI-level grounding. Our dataset comprises 600 samples from 200 patients with 1,960 RoIs and reports in Vietnamese. Each RoI is manually labeled with comprehensive clinical attributes, providing a structured foundation for learning from segmentation to generation. 

Furthermore, we propose \textit{HiRRA}, a novel VLM architecture designed to mimic the professional diagnostic workflow. \textit{HiRRA} comprises two core components: a dual-stream encoder and a graph-based relational module. The former preserves separate CT and PET features before fusion to ensure high-fidelity volumetric data; the latter models inter-RoI dependencies vital for diagnosing metastatic patterns and systemic disease \citep{ju2024graph, kazmierski2021lymph}. Experimental results demonstrate that HiRRA substantially outperforms existing baselines, achieving a 19.7\% improvement in BLEU-4 over traditional end-to-end methods. To ensure a more rigorous evaluation of clinical utility, we also propose two clinical metrics: RoI Coverage and RoI Quality Index, which measure the fidelity of region-level attribute descriptions. 

In summary, the primary contributions of our work include:

\vspace{-0.1cm}
\begin{itemize}\itemsep2pt\parskip0pt\topsep5pt
\item First, we introduce \textit{VietPET-RoI}, the first public RoI-grounded PET/CT dataset with 600 samples from 200 patients and 1,960 RoIs with comprehensive clinical annotations, addressing RoI-level supervision gap and medical AI scarcity for low-resource languages.

\item Second, we propose \textit{HiRRA}, a 
VLM emulating the 
diagnostic workflow. HiRRA integrates CT and PET information through hierarchical feature extraction, capturing global volumetric and localized RoI features, achieving SOTA on linguistic and clinical metrics.

\item Third, we design clinical metrics specifically tailored for medical report generation, directly assessing clinical factors such as \textit{RoI Coverage} and \textit{RoI Quality Index}.

\end{itemize}

\vspace{-0.4cm}
\section{Related work}
\subsection{Existing Medical 3D datasets}

\begin{figure*}[t] 
    \centering
    \vspace{-0.1cm}
    \includegraphics[width=0.97\textwidth]{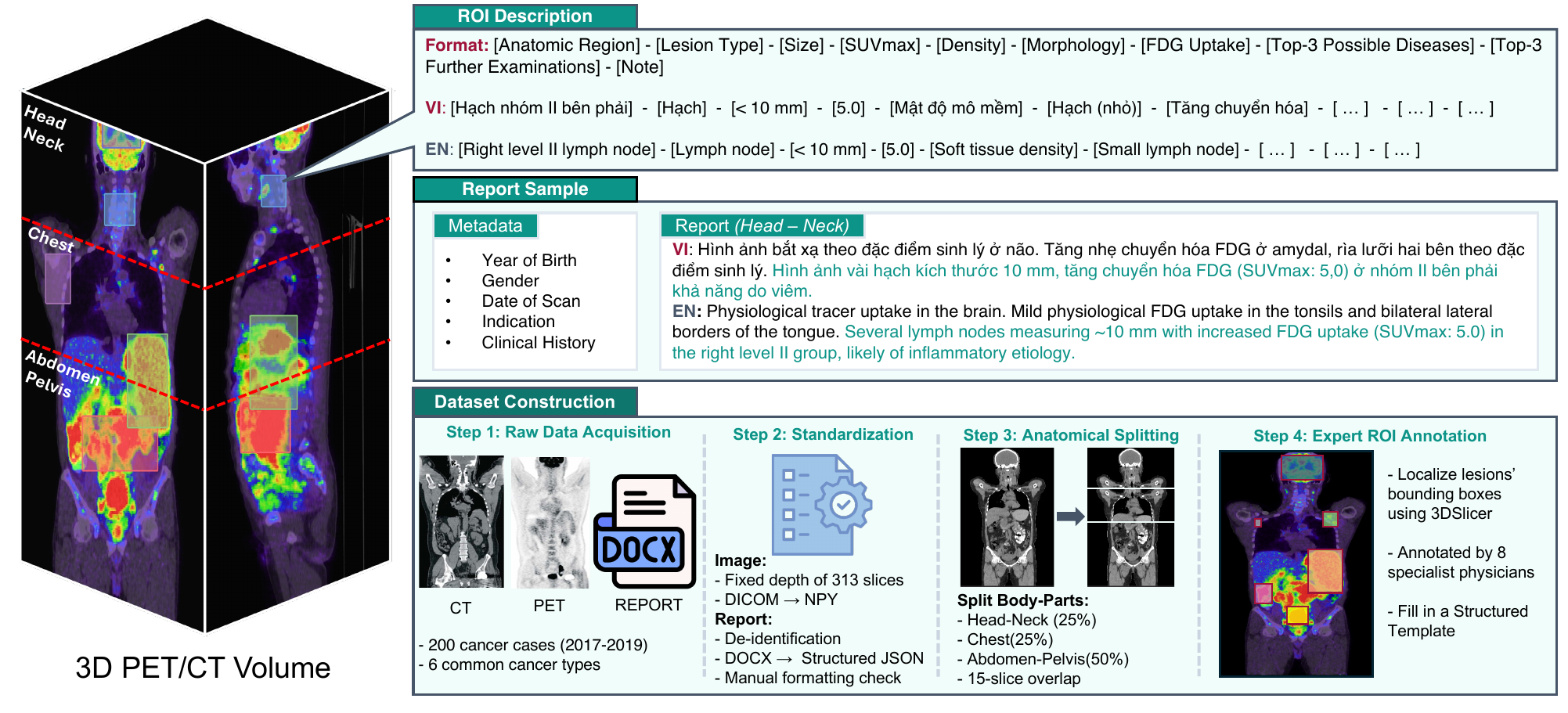} 
    \vspace{-0.3cm}
    \caption{\textbf{Overview of the VietPET-RoI dataset.} The figure displays (top) the multimodal data samples including 3D PET/CT volumes, structured RoI descriptions, and clinical reports; and (bottom) the four-stage curation pipeline, spanning from raw data acquisition to expert-level annotation.}
    \label{fig:dataset_visualize}
    \vspace{-0.5cm}
    
\end{figure*}
While large-scale PET/CT datasets like AutoPET~\cite{gatidis2022whole} and RIDER~\cite{muzi2015data} support dense segmentation or lesion detection, they lack aligned clinical reports, limiting their utility for multimodal modeling. Conversely, recent volumetric report generation benchmarks, including ViMed-PET~\cite{nguyen2026toward}, PETRG-Lym~\cite{jiao2025vision}, and PET2Rep~\cite{zhang2026pet2rep}, pair scans with diagnostic text but rely on coarse-grained supervision, omitting explicit lesion localization. This dichotomy highlights a critical gap: the lack of datasets combining fine-grained region grounding with diagnostic reporting to enable interpretable, clinically aligned modeling.
\vspace{-0.2cm}

\subsection{3D Report Generation VLMs}

Recent advances in vision–language models (VLMs) have extended medical report generation to 3D volumetric imaging, enabling richer spatial reasoning over CT, MRI, and PET/CT scans. Representative models like M3D~\cite{bai2024m3d} and Med3DVLM~\cite{xin2025med3dvlm} have integrated 3D encoders with large language models, achieving promising results when generating global diagnostic reports from full volumes. However, these end-to-end approaches rely on scan-level supervision, lacking explicit region-level grounding. This diverges from standard clinical practice, where radiologists follow a region-centric workflow - detecting, localizing, and characterizing abnormalities before report synthesis. Consequently, existing models fail to capture this intermediate reasoning, creating a gap between volumetric perception and clinically meaningful, structured generation.

\section{VietPET-RoI: A RoI-Grounded Vietnamese 3D PET/CT Dataset}
We detail the dataset’s pipeline and clinical statistics, proving its suitability for 3D report generation.
\vspace{-0.4cm}
\subsection{Dataset Construction Process}

The dataset was curated by oncology specialists at a leading public hospital in Vietnam, comprising 200 cancer cases spanning the six most common malignancies examined between 2017 and 2019. PET/CT scans were standardized to a unified format with 313 axial slices from head to upper thighs, then divided into three anatomical regions (head-neck, chest, abdomen-pelvis) with 15-slice overlap to yield 600 region-level image-report pairs (shown in Figure~\ref{fig:dataset_visualize}). Details on preprocessing procedures (DICOM conversion, de-identification, report standardization) are provided in Appendix ~\ref{sec:appendix_dataset}.

The key novelty of \textit{VietPET-RoI} lies in fine-grained RoI annotations. Unlike existing PET/CT datasets that provide only global scan-report pairs, eight nuclear medicine physicians collaborated to localize and describe every abnormal finding or clinically relevant physiological structure mentioned in the reports. Using 3DSlicer~\cite{pieper20043d}, physicians marked each RoI with a 3D bounding box and completed a structured template recording ten clinical attributes: anatomical location, lesion type, size, SUVmax, density, morphology, FDG uptake, top-3 differential diagnoses with follow-up examinations, and additional notes.

A total of 1,960 RoIs were annotated, with all samples rigorously verified by senior nuclear medicine physicians to ensure consistency. Any discrepancies were resolved through consensus discussion to guarantee high-quality ground truth. 

Unlike existing datasets providing only the final reports, our fine-grained RoI annotations enable models to learn direct associations between specific regions and corresponding clinical findings, rather than relying on ambiguous global image-text mappings. Moreover, this design replicates the clinical diagnostic workflow, facilitating multi-task learning including segmentation, description generation, and decision support.
\vspace{-0.1cm}

\subsection{Dataset Characteristics}

\begin{figure}[t]
    \centering
    \includegraphics[width=\linewidth]{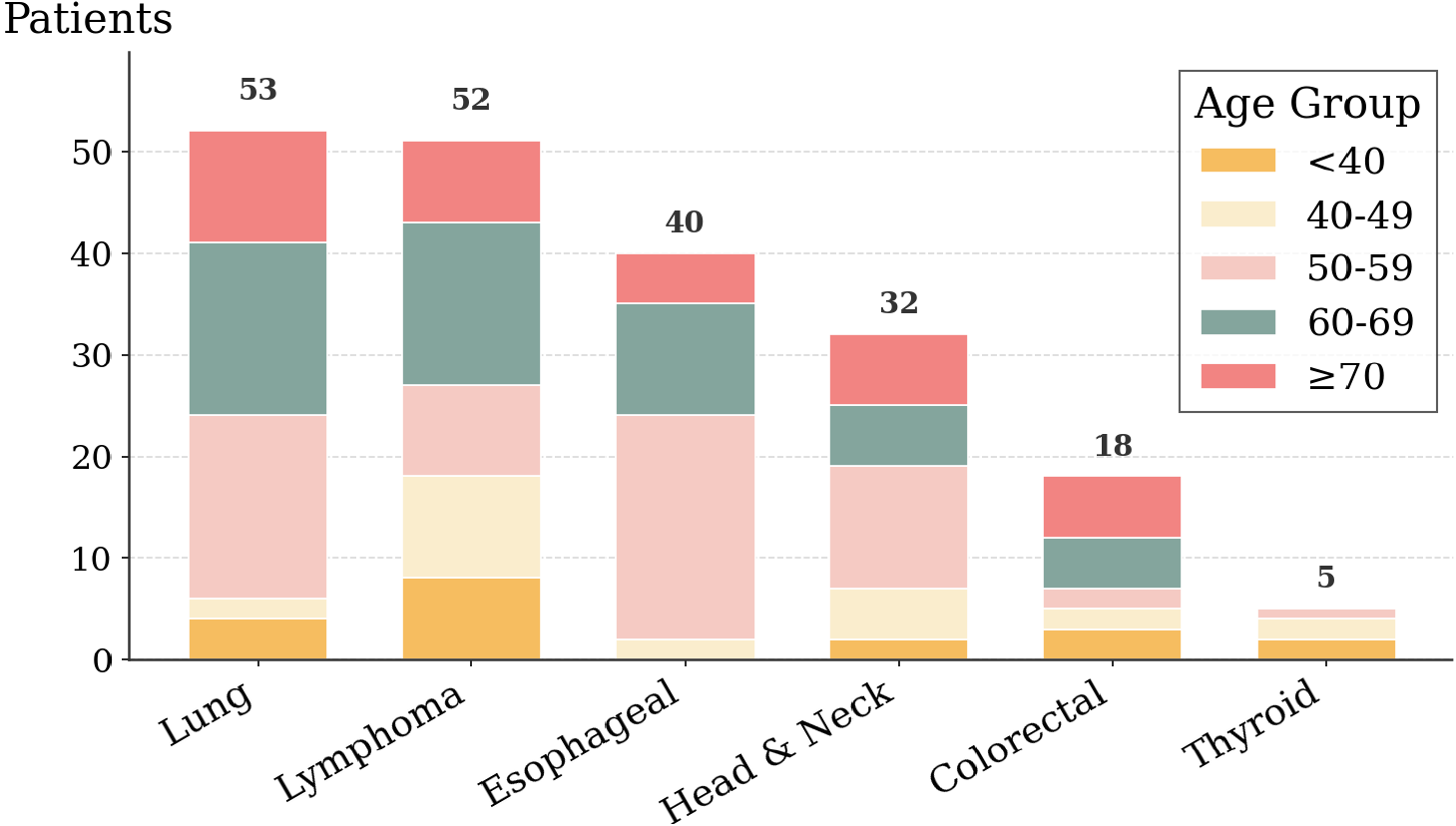}
    \caption{Data distribution across the six cancer types.}
    \label{fig:cancer_gender}
    \vspace{-0.4cm}
\end{figure}

\textit{VietPET-RoI} provides a multimodal dataset comprising textual reports, 3D PET/CT volumes, and fine-grained RoI-description. In total, the dataset contains 600 region-level samples derived from 200 cancer patients covering six common and clinically important malignancies: lymphoma, head and neck cancer, lung cancer, esophageal cancer, thyroid cancer and colorectal cancer. The cohort was selected to include diverse sex, age, and body habitus based on patient metadata, reducing bias toward any specific demographic group. Figure~\ref{fig:cancer_gender} shows the age distribution across the six cancer types, while key dataset-level statistics are summarized in Table~\ref{tab:dataset_summary}.

For each study, the CT volumes are stored as 3D arrays of size \(\text{slices} \times 512 \times 512\), whereas the PET volumes are represented as \(\text{slices} \times 256 \times 256\), following typical clinical in-plane resolutions and allowing straightforward voxel-wise fusion between modalities. After RoI annotation, we observe an average of 3.27 RoIs per body-region sample, with a minimum of 1 and a maximum of 23 RoIs. The abdomen\_pelvis region is the most densely annotated, typically containing  approximately 3.9 RoIs per sample, reflecting the prevalence of abdominal and pelvic lesions in the target cancer types. Approximately 42\% of RoIs correspond to pathological findings, while about 58\% capture physiological uptake patterns and serve as negative or contextual examples for the models. This configuration yields a compact but clinically rich dataset with dense region-level supervision, well suited for training and evaluating RoI-grounded PET/CT report generation and other multimodal medical AI tasks. The dataset will be publicly released for non-commercial research purposes under appropriate usage agreements upon paper acceptance.
\vspace{-0.1cm}

\section{HiRRA: A Hierarchical Region-Aware Framework for Report Autogeneration}

\begin{figure*}[ht]
    \centering
    \includegraphics[width=0.95\textwidth]{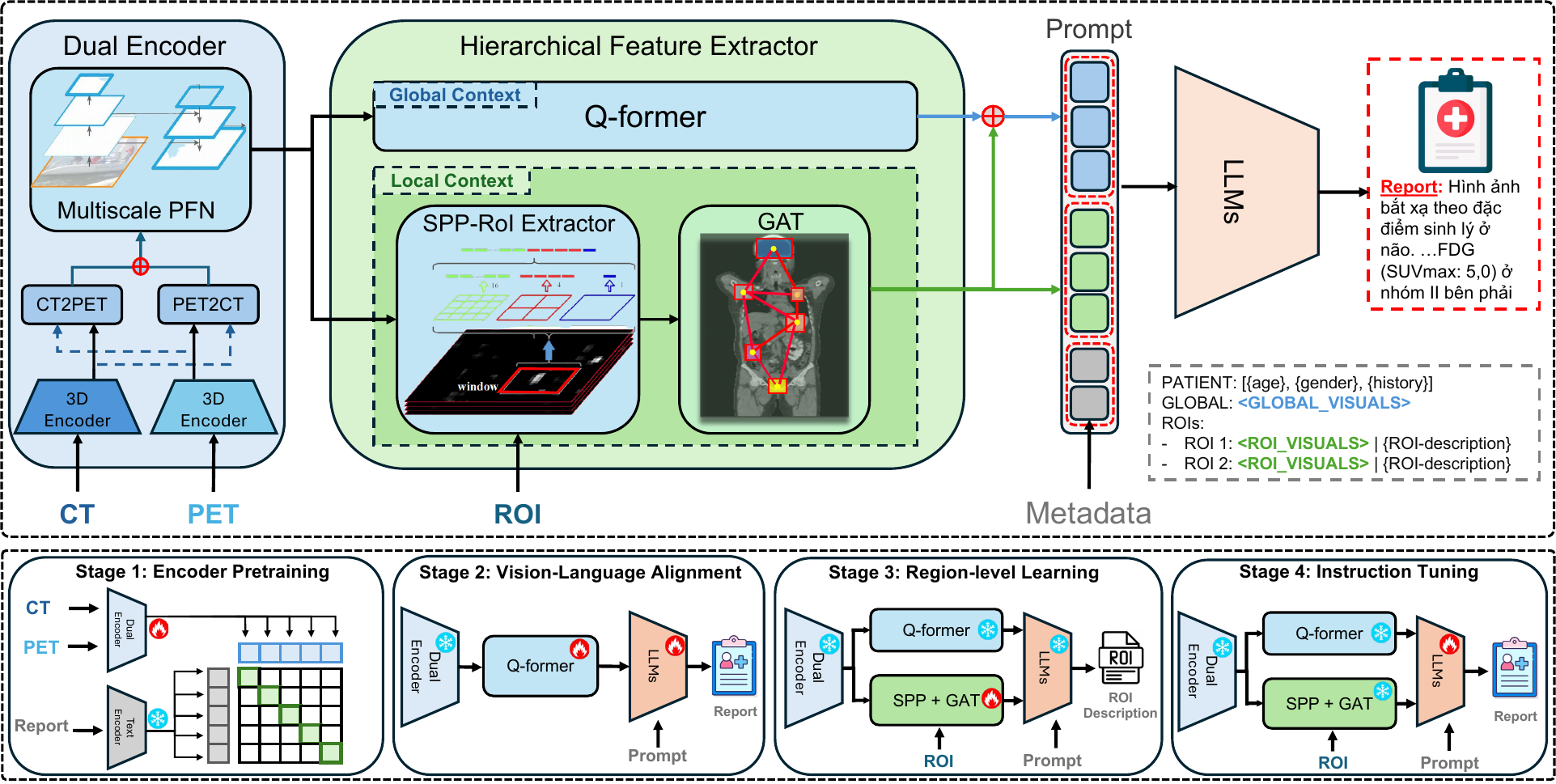} 
    \caption{\textbf{The overall architecture of HiRRA.} The framework processes paired PET/CT volumes through a Dual Encoder and a Hierarchical Feature Extractor. The \textit{Global Context} is captured via Q-former, while the \textit{Local Context} is using SPP-RoI extraction and GATv2. Finally, the LLM generates the report using a semantic-injected prompt.}
    \label{fig:HiRRA_framework}
    \vspace{-0.4cm}
\end{figure*}

\begin{table}[t]
    \centering
    \caption{Key characteristics of the VietPET-RoI dataset.}
    \vspace{-0.1cm}
    \label{tab:dataset_summary}
    \small
    \setlength{\tabcolsep}{8pt}
    \renewcommand{\arraystretch}{1.15}
    \resizebox{\columnwidth}{!}{%
    \begin{tabular}{l l}
        \toprule
        \textbf{Characteristic} & \textbf{Value} \\
        \midrule
        Patients & 200 \\
        Region-level samples & 600 \\
        Age (mean $\pm$ SD) & $57.9 \pm 12.2$ years \\
        Age range & 13--81 years \\
        CT volume size & $313 \times 512 \times 512$ \\
        PET volume size & $313 \times 256 \times 256$ \\
        Total RoIs & 1,960 \\
        Avg RoIs/sample & 3.27 \\
        Most annotated region & Abdomen-pelvis ($\approx$3.9/sample) \\
        RoI composition & 42\% pathological / 58\% physiological \\
        \bottomrule
    \end{tabular}%
    }
    \vspace{-0.6cm}
\end{table}
\noindent \textbf{Overview.} 
To empirically validate the utility of the VietPET-RoI dataset, we propose HiRRA, a Hierarchical Region-aware Framework for Report Autogeneration. Unlike conventional end-to-end VLMs that directly map global image features to textual reports, HiRRA is architected to emulate the professional diagnostic workflow, where physicians synthesize reports by integrating holistic volumetric scans with a granular analysis of specific RoIs. As illustrated in Figure~\ref{fig:HiRRA_framework}, the HiRRA framework is composed of three primary modules designed to facilitate this hierarchical translation.
The first component is the Dual Encoder, which independently processes anatomical CT and functional PET to preserve high-fidelity features prior to fusion. Subsequently, a Hierarchical Feature Extractor utilizes a bifurcated strategy: a Global Context block captures overarching volumetric characteristics, while a Local Context block extracts fine-grained features from annotated RoIs. Finally, an LLM-based Decoder integrates these multi-granularity features to generate clinical reports. By guiding attention through RoI visual tokens, this structured supervision ensures the model significantly outperforms traditional methods trained only on image-text pairs.

\noindent\textbf{Multimodal Dual Encoder.} Motivated by the clinical workflow where physicians overlay metabolic heatmaps (PET) onto anatomical scans (CT), we design a dual-stream architecture to extract modality-specific features before fusion. Given paired volumes $I_{CT}, I_{PET} \in \mathbb{R}^{B \times 1 \times T \times H \times W}$, we employ two separate 3D encoders $\mathcal{E}_{CT}$ and $\mathcal{E}_{PET}$ based on CT-ViT~\cite{hamamci2024ct2repautomatedradiologyreport} to extract features $F_{CT}, F_{PET} \in \mathbb{R}^{B \times N \times D}$, where $N = T' \times H' \times W'$ and $D=512$. Unlike single-encoder approaches that prematurely fuse CT and PET channels, our dual-stream design preserves modality-specific characteristics: CT captures anatomical structures (tissue boundaries, organ morphology) while PET quantifies metabolic activity (glucose uptake patterns)~\cite{beyer2004acquisition}.

To bridge the semantic gap, we apply bidirectional cross-attention~\cite{li2023blip}:
\vspace{-0.1cm}
\begin{align}
\tilde{F}_{CT} &= F_{CT} + \text{CA}(F_{CT}, F_{PET}), \\
\tilde{F}_{PET} &= F_{PET} + \text{CA}(F_{PET}, F_{CT}),
\end{align}
\vspace{-0.1cm}
where $\text{CA}(\cdot, \cdot)$ denotes cross-attention with query from the first argument and key-value from the second. The fused representation $F_{visual}$ is obtained by averaging $\tilde{F}_{CT}$ and $\tilde{F}_{PET}$, followed by a Multiscale Pyramid Feature Network (PFN)~\cite{lin2017featurepyramidnetworksobject} to capture multi-resolution features.

\noindent\textbf{Hierarchical Feature Extractor.} 
To synthesize comprehensive visual representations, we extract features at two distinct levels of granularity: global and local context. The Global Context block captures the overarching environment from the input CT and PET volumes. This is operationalized using a Q-former~\cite{li2023blip} with query vectors, which compresses the high-dimensional volumetric data into a fixed-length latent representation.

For the Local RoI-Aware Context, we extract region-specific information to augment the LLM’s inferential capabilities during report generation. Traditional methods typically process RoIs independently, thereby overlooking the critical clinical relationships between disparate lesions. However, diagnostic accuracy in PET/CT often hinges on these relationships: spatially proximate lesions may indicate local invasion, while distant lesions exhibiting similar metabolic patterns frequently suggest metastatic spread. To effectively capture these dependencies, we employ a Graph Neural Network to learn the spatio-morphological correlations between RoIs. Specifically, we construct a graph whose each node represents an individual RoI. Given $N$ RoIs with features $\mathbf{h}_i \in \mathbb{R}^D$ and bounding boxes $\mathbf{b}_i \in \mathbb{R}^6$, we establish edges based on two criteria: spatial proximity (geometric distance $d_{ij}$ between centroids $\mathbf{c}_i$ and $\mathbf{c}_j$) and morphological similarity (feature cosine similarity $s_{ij}$). An edge $(i,j)$ is created if $d_{ij} < \tau_d$ or $s_{ij} > \tau_s$, enabling the graph to capture both adjacent lesions and distant metastases.

For each edge, we encode the relationship by constructing edge features:
\vspace{-0.1cm}
\begin{equation}
\mathbf{e}_{ij} = \text{MLP}\big([\mathbf{h}_i \| \mathbf{h}_j \| \mathbf{g}_{ij}^s \| \mathbf{g}_{ij}^m]\big),
\end{equation}
where $\mathbf{g}_{ij}^s = [d_{ij}, \mathbf{r}_{ij}, v_i/v_j]$ encodes spatial features (distance, relative direction, volume ratio), $\mathbf{g}_{ij}^m = [s_{ij}, \bar{I}_i, \bar{I}_j]$ encodes morphological features (similarity and mean intensities $\bar{I}$ from CT/PET at RoI locations), and $\|$ denotes concatenation.
We apply GATv2~\cite{brody2022attentivegraphattentionnetworks} with its default attention mechanism for message passing, producing enhanced representations $\mathbf{h}_i'$ that incorporate contextual information from spatially adjacent and morphologically similar lesions.

\noindent\textbf{Semantic-Injected LLM Decoder.} Instead of relying solely on visual tokens, HiRRA employs a description-guided prompting strategy. The LLM decoder input consists of a composite prompt: (1) patient demographics and clinical history; (2) global visual tokens \texttt{<GLOBAL\_VISUALS>}; and (3) structured RoI information \texttt{RoI $i$: <RoI\_VISUALS> | \{RoI-description\}}, where \texttt{<RoI\_VISUALS>} represents projected enhanced features $\mathbf{h}_i'$ and \texttt{\{RoI-description\}} contains structured clinical attributes (SUVmax, size, FDG).

\noindent\textbf{Training Strategy.} We implement a four-stage training paradigm, wherein tailored prompting schemes are designed for each phase to facilitate progressive alignment and learning.

\noindent \textit{Stage 1: Encoder Pretraining.} We pretrain dual encoders on ViMed-PET~\cite{nguyen2026toward} using CLIP-style contrastive learning, aligning CT-PET features with Vietnamese text reports.

\noindent \textit{Stage 2: Vision-Language Alignment.} With encoders and LLM frozen, we train only the Q-former and projection layers. The prompt contains only patient metadata and \texttt{<GLOBAL\_VISUALS>} tokens for image-report alignment.

\noindent \textit{Stage 3: Region-level Learning.} We integrate local context modules (SPP-RoI, GATv2), freezing Stage 2 components. The prompt extends to full structure with \texttt{<RoI\_VISUALS>} tokens and RoI description slots. Using bounding box supervision, the model generates RoI-level attribute descriptions.

\noindent \textit{Stage 4: Instruction Tuning.} We fine-tune end-to-end via LoRA~\cite{hu2022lora} ($r=16$, $\alpha=32$) with unfrozen RoI vision modules. Using the full prompt template, the model synthesizes RoI findings into complete Vietnamese reports.
\vspace{-0.2cm}

\section{Evaluation}
\subsection{Evaluation Metrics}

To ensure a comprehensive assessment of both linguistic fluency and clinical validity, we employ a dual-evaluation strategy comprising standard linguistic metrics and a proposed clinical protocol.
\paragraph{Standard NLP Metrics.}
We report n-gram and embedding-based metrics. i.e., BLEU, ROUGE, BERTScore  to evaluate lexical and semantic similarity. All scores are reported as percentages. 
\paragraph{Proposed Clinical Evaluation Protocol.}
While the standard linguistic metrics effectively measure the text overlap, they often fail to capture the clinical correctness, anatomical precision, and diagnostic hierarchy essential for 3D PET/CT reporting. To this end, we introduce a structured protocol assessing both the correctness and semantic fidelity of the RoIs identified in the generated reports. 
To enable this, we first utilize \textit{LangExtract}~\cite{Goel_LangExtract_2025} to parse both ground truth and generated reports into structured objects defined by five key clinical fields: $E=\{\text{region, lesion, density, morphology, fdg\_uptake}\}$. Based on this structured representation, we define two clinical metrics as follows. 

\noindent \textbf{\textit{RoI Coverage}} - \textit{Quantitative Identification}. 
We frame RoI evaluation as a bipartite matching problem. Specifically, we compute a pairwise similarity matrix between the set of predicted RoI text spans $\hat{\mathcal{R}}$ and ground-truth spans $\mathcal{R}$, utilizing cosine similarity within a BERT-based embedding space. Optimal one-to-one alignment is established via the Hungarian matching algorithm~\cite{kuhn1955hungarian}.

Pairs exceeding similarity threshold $\tau$ are defined as True Positives ($TP$), forming the basis for Precision, Recall, and F1-score calculations.\\

\noindent \textbf{\textit{RoI Quality Index}} - \textit{Qualitative Fidelity}. 
For matched RoI pairs, we evaluate attribute-specific accuracy using the proposed RoI Quality Index (RoIQ). This metric is designed to enforce a clinical hierarchy, operationalizing the principle that accurate anatomical region and lesion type identification are fundamental prerequisites for a valid diagnostic description. To reflect this, we define $S_{\text{region}}$ and $S_{\text{lesion}}$ as the similarity scores for these critical attributes, while $\mathcal{A}_{\text{valid}}$ represents the set of additional non-empty attributes extracted from the clinical fields. The RoIQ is formulated as follows:
\vspace{-0.1cm}
\begin{equation} \nonumber
    \text{RoIQ} = \sqrt{S_{\text{region}} \cdot S_{\text{lesion}}} \times \left( \frac{1}{|\mathcal{A}_{\text{valid}}|} \sum_{k \in \mathcal{A}_{\text{valid}}} S_{k} \right).
\end{equation}
\vspace{-0.1cm}
The first term, represented by the geometric mean of the core attributes, serves as a non-linear penalty for hallucinations in critical anatomical or pathological contexts. By structuring the metric in this manner, we ensure that high performance in secondary descriptors (e.g., size or morphology) cannot compensate for fundamental errors in localization or lesion categorization. Consequently, RoIQ provides a more clinically grounded assessment of report quality than standard token-based overlaps, emphasizing diagnostic reliability.
\\
Our metrics were informed by the clinical intuition of physicians from a hospital in Vietnam, and their medical correctness and clinical usefulness were subsequently validated by these clinicians.
\vspace{-0.2cm}

\subsection{Experimental Goals}

To comprehensively evaluate the VietPET-RoI benchmark and the proposed HiRRA framework, we investigate the following research questions:

\noindent\label{rq:lowresource} \textbf{RQ1 - Performance on Low-Resource Languages:} We benchmark representative VLMs on VietPET-RoI to assess their capability in handling native Vietnamese medical reports versus translated data in English versions.

\noindent \textbf{RQ2 - Hallucination Mitigation and RoI Grounding:\label{rq:hallucination} } We evaluate the efficacy of the HiRRA pipeline compared to fine-tuned baselines in reducing hallucinations. This involves our new metric that measures both the quantitative recall of detected regions and the qualitative semantic accuracy of their descriptions.
    
\noindent \textbf{RQ3 - Performance on Clinical Tasks:\label{rq:clinical}} We assess model performance on clinical tasks, specifically differential diagnosis prediction and RoI description generation, to demonstrate dataset versatility.
    
\noindent \textbf{RQ4 - Impact of Multimodal Fusion:\label{rq:fusion} } We compare single-modality (CT or PET only) versus joint CT+PET fusion to assess the necessity of integrating anatomical and metabolic information for accurate RoI characterization.
\vspace{-0.35cm}

\subsection{Model selection}

To establish a robust benchmark for \textit{VietPET-RoI}, we evaluate representative state-of-the-art 2D and 3D vision-language models. For 2D modeling, we compare \textit{InternVL3}~\cite{chen2024internvl,zhu2025internvl3}, a large-scale generalist foundation model, against \textit{MedGemma}~\cite{sellergren2026medgemmatechnicalreport}, a specialized medical backbone. For 3D volumetric reasoning, we incorporate four leading medical VLMs: \textit{RadFM}~\cite{wu2025towards} and \textit{MedM-VL}~\cite{shi2025medm} as general-purpose backbones with large-scale multimodal pretraining, and \textit{Med3DVLM}~\cite{xin2025med3dvlm} and \textit{M3D-LaMed}~\cite{bai2024m3d} with architectures optimized for localization and report generation. While models like \textit{CT2Rep}~\cite{hamamci2024ct2repautomatedradiologyreport} and \textit{PETAR}~\cite{maqbool2025petarlocalizedfindingsgeneration} advance report generation, they have distinct structural limitations: CT2Rep is restricted to global, CT-only volume-to-report mapping, and PETAR requires pre-existing oracle lesion masks to generate isolated lesion captions. In contrast, our framework uniquely performs end-to-end RoI discovery, structured multi-attribute modeling, and explicit inter-RoI relational reasoning via GATv2 to synthesize coherent whole-body PET/CT reports

\subsection{Performance on Low-Resource Languages}
\begin{table}[t!]
\centering
\caption{\textbf{Benchmarking VLMs on VietPET-RoI.} Results are reported for Vietnamese (VI) and English (EN) inference. Best scores for VI are in \textbf{bold} and \underline{underline} for EN. All metrics are reported as percentages (\%).}

\label{tab:rq1_results}

\resizebox{0.92\linewidth}{!}{%
\begin{tabular}{l c c c c c}
\toprule
\textbf{Model} & \textbf{Lang} & \textbf{BLEU-4} & \textbf{R-1} & \textbf{R-L} & \textbf{BERT} \\
\midrule

\multirow{2}{*}{InternVL3 (2D)} & VI & 0.25 & \textbf{28.20} & \textbf{18.28} & 64.05 \\
                               & EN & \underline{0.94} & 16.64 & 11.05 & 81.50 \\
\midrule

\multirow{2}{*}{MedGemma (2D)} & VI & \textbf{0.82} & 25.59 & 17.61 & 65.36 \\
                               & EN & 0.39 & 13.58 & 8.87 & 80.12 \\
\midrule

\multirow{2}{*}{RadFM}         & VI & 0.39 & 19.09 & 13.84 & \textbf{83.33} \\
                               & EN & 0.44 & 12.32 & 8.78 & 80.68 \\
\midrule

\multirow{2}{*}{MedM-VL}       & VI & 0.13 & 1.41 & 1.36 & 47.28 \\
                               & EN & 0.93 & \underline{22.02} & \underline{15.15} & \underline{84.27} \\
\midrule

\multirow{2}{*}{Med3DVLM}      & VI & 0.17 & 1.31 & 1.24 & 62.67 \\
                               & EN & 0.25 & 12.54 & 9.44 & 67.14 \\
\midrule

\multirow{2}{*}{M3D-LaMed}     & VI & 0.15 & 1.18 & 1.08 & 62.92 \\
                               & EN & 0.45 & 13.42 & 9.96 & 67.44 \\

\bottomrule
\end{tabular}%
}
\vspace{-0.6cm}
\end{table}

\begin{table*}[htb]
\centering
\caption{\textbf{Quantitative benchmarking against state-of-the-art methods.} We evaluate the generation quality (NLP Metrics) and clinical alignment (RoI Clinical Metrics). \textit{Correct} indicates the number of successfully grounded regions (out of 416). \textbf{Bold} denotes the best performance. The rows highlighted in gray show our model's performance and its relative improvement ($\uparrow$) over the \underline{second-best} baseline.}
\label{tab:main_results}

\begin{adjustbox}{width=0.95\textwidth}
\begin{tabular}{l cccc cccccc}
\toprule
\multirow{2}{*}{\textbf{Method}} & 
\multicolumn{4}{c}{\textbf{Generation Quality (NLP)}} & 
\multicolumn{6}{c}{\textbf{Clinical RoI Alignment}} \\ 
\cmidrule(lr){2-5} \cmidrule(lr){6-11}

& \textbf{BLEU-4} & \textbf{R-1} & \textbf{R-L} & \textbf{BERT} & 
\textbf{Correct} & \textbf{Total} & \textbf{Prec.} & \textbf{Rec.} & \textbf{F1} & \textbf{RoIQ} \\ 
\midrule

MedM-VL   & 31.69 & 67.11 & 50.00 & 91.92 & 62  & 416 & 12.88 & 19.12 & 14.16 & 23.24 \\
Med3D-VLM  & 45.53 & 71.06 & 62.51 & 86.49 & 179 & 416 & 31.07 & 43.02 & 36.08 & \underline{39.00} \\
M3D-LaMed & 44.30 & 74.14 & 64.39 & 85.90 & \underline{193} & 416 & \underline{34.90} & \underline{46.39} & \underline{39.83} & 36.31 \\

\midrule

HiRRA (No RoI) & \underline{52.48} & \underline{77.75} & \underline{66.51} & \underline{95.13} & 187 & 416 & 32.29 & 44.95 & 37.58 & 33.89 \\

\rowcolor{gray!10} 
\textbf{HiRRA (Ours)} & 
\textbf{62.80} & \textbf{80.40} & \textbf{69.66} & \textbf{95.79} & 
\textbf{223} & 416 & 
\textbf{35.17} & \textbf{53.60} & \textbf{42.47} & \textbf{56.86} \\ 

\textit{Improv. ($\Delta\%$)} & 
\inc{+19.7\%} & \inc{+3.4\%} & \inc{+4.7\%} & \inc{+0.7\%} & 
\inc{+15.5\%} & - & 
\inc{+0.8\%} & \inc{+15.5\%} & \inc{+6.6\%} & \textbf{\textcolor{ForestGreen}{\footnotesize +45.8\%}} \\

\bottomrule
\end{tabular}
\end{adjustbox}
\vspace{-0.2cm}
\end{table*}
To evaluate the capability of existing medical VLMs on low-resource languages as proposed in \hyperref[rq:lowresource]{RQ1}, we benchmark state-of-the-art 2D and 3D models on VietPET-RoI. Results in Table~\ref{tab:rq1_results} reveal that all models perform extremely poorly on Vietnamese generation, with BLEU-4 scores near zero (MedGemma: 0.82, Med3D-VLM: 0.17, M3D-LaMed: 0.15). This collapse arises from a pronounced mismatch between existing model pretraining, which primarily covers English radiology text and 2D or 3D CT volumes, and the VietPET-RoI setting that requires reasoning over Vietnamese clinical language, 3D PET/CT volumes, and structured region-level clinical descriptions. In addition, VietPET-RoI requires models to describe metabolic activity (SUV$_{\text{max}}$), lesion morphology, and precise anatomical regions, making it a challenging benchmark that tests true PET/CT understanding rather than surface-level text generation.

\subsection{Hallucination Mitigation and RoI Grounding}
\begin{table}[tbh]
\centering
\caption{\textbf{Clinical Prediction and RoI Description Generation performance (\hyperref[rq:clinical]{RQ3}}). We report Recall (Rec.) and Precision(Prec.) for clinical tasks, and standard NLP metrics (BLEU-4,  ROUGE-L) for description generation. Best results are highlighted in bold.}
\label{tab:rq3_stacked}
\resizebox{0.96\linewidth}{!}{%
\begin{tabular}{l@{\hspace{8pt}}cc@{\hspace{8pt}}cc@{\hspace{8pt}}cc}
\toprule
\multirow{2}{*}{\textbf{Model}} & 
\multicolumn{2}{c@{\hspace{8pt}}}{\textbf{Disease Pred.}} & 
\multicolumn{2}{c@{\hspace{8pt}}}{\textbf{Exam. Pred.}} & 
\multicolumn{2}{c}{\textbf{RoI Desc.}} \\
\cmidrule(lr){2-3} \cmidrule(lr){4-5} \cmidrule(lr){6-7}
& \textbf{Prec.} & \textbf{Rec.} & 
\textbf{Prec.} & \textbf{Rec.} & 
\textbf{B-4} & \textbf{R-L} \\
\midrule
M3D-LaMed & 
67.4 & 73.2 & 
21.6 & 21.3 & 
36.0 & 69.3 \\
Med3D-VLM & 
76.5 & 79.9 & 
41.0 & 42.3 & 
\textbf{62.3} & 75.8 \\
MedM-VL & 
\textbf{95.0} & \textbf{98.7} & 
\textbf{51.7} & \textbf{52.0} & 
50.3 & \textbf{78.8} \\
\bottomrule
\end{tabular}%
}
\vspace{-0.5cm}
\end{table}
To address \hyperref[rq:hallucination]{RQ2}
, we compare HiRRA against three top-performing 3D VLMs (MedM-VL~\cite{shi2025medm}, Med3DVLM~\cite{xin2025med3dvlm}, M3D-LaMed~\cite{bai2024m3d}) fully fine-tuned on VietPET-RoI. Additionally, we evaluate HiRRA under two configurations: \textit{No RoI} (global context only) and \textit{Full Pipeline} (with RoI features). Evaluation employs both traditional NLP metrics and our proposed clinical metrics: \textit{RoI Coverage} and \textit{RoI Quality Index}.

Results in Table~\ref{tab:main_results} show that HiRRA Full Pipeline achieves state-of-the-art performance across all metrics. For generation quality, HiRRA attains BLEU-4 of 62.80 and BERTScore of 95.79, substantially outperforming the strongest baseline (Med3DVLM). More importantly, HiRRA demonstrates significant improvements in clinical alignment with 45.8\% gain in RoIQ and 15.5\% increase in RoI Recall, indicating superior hallucination reduction. Ablation comparison confirms the critical role of RoI: \textit{HiRRA No RoI} maintains strong fluency (BERT 95.13) but achieves only 0.3389 in RoIQ, proving that global context alone is insufficient for clinical accuracy. This validates that RoI grounding is essential for bridging visual perception with accurate diagnostic reporting.
\vspace{-0.1cm}

\subsection{Performance on Clinical Tasks}
We assess whether VietPET-RoI supports multitask learning by training models on two tasks: differential diagnosis and further examinations prediction, structured RoI description generation. Results are summarized in Table~\ref{tab:rq3_stacked}.

For clinical prediction, MedM-VL~\cite{shi2025medm} achieves the highest Recall on disease prediction (98.7\%), while examination prediction remains significantly lower (52.0\%), indicating that follow-up recommendation requires more complex reasoning. For RoI description generation, models achieve strong performance with BLEU-4 up to 62.3\% and ROUGE-L of 78.8\%, demonstrating that VietPET-RoI provides effective fine-grained supervision for generation tasks. These results confirm that VietPET-RoI supports multitask learning, creating potential for a complete automated pipeline: from RoI segmentation to region-specific description generation, faithfully mirroring clinical workflows and paving the way for future diagnostic support systems.

\subsection{Impact of Multimodal Fusion}
Unlike prior approaches often limited to single-modality inputs (e.g., ViMed-PET~\cite{nguyen2026toward}), Table~\ref{tab:rq4_fusion} confirms the necessity of multimodal fusion. The joint CT+PET framework achieves superior performance (36.1 BLEU-4), significantly outperforming single-modality baselines by a margin of 2.6--4.9 points. This validates that integrating anatomical structure (CT)~\cite{kinahan2006petct} with metabolic intensity (PET)~\cite{townsend2008petct, von2009petct} is indispensable for accurate lesion description, effectively resolving the perceptual bottlenecks inherent in unimodal processing. Specifically, this cross-modal synergy enables the disambiguation of hypermetabolic signals, allowing the model to distinguish between pathological lesions and physiological uptake (e.g., in the brain or bladder) through anatomical referencing, which significantly enhances the report's clinical validity.
\begin{table}[t]
\centering
\caption{\textbf{Impact of Multimodal Fusion.} Quantitative ablation demonstrating the superiority of the joint CT+PET framework over single-modality baselines.}
\label{tab:rq4_fusion}

\setlength{\tabcolsep}{4pt}

\resizebox{0.9\linewidth}{!}{%
\begin{tabular}{l c c c c}
\toprule
\textbf{Input Modality} & \textbf{BLEU-4} & \textbf{R-1} & \textbf{R-L} & \textbf{BERT} \\
\midrule

CT only & 31.3 & 65.4 & 56.3 & 92.7 \\

PET only & 33.6 & 66.3 & 58.0 & 93.1 \\

\rowcolor{gray!15} 
\textbf{CT + PET} & \textbf{36.1} & \textbf{69.3} & \textbf{59.4} & \textbf{94.0} \\

\bottomrule
\end{tabular}%
}
\vspace{-0.4cm}
\end{table}
\vspace{-0.2cm}

\section{Conclusion}

We addressed the scarcity of fine-grained PET/CT datasets 
by introducing a novel dataset that comprises 600 region-level samples with 1,960 comprehensively annotated RoIs, representing the first 3D PET/CT dataset with fine-grained RoI-level annotations. We also proposed HiRRA, a novel VLM architecture that integrates local RoI features and global context for report generation. Additionally, we introduced clinical evaluation metrics (RoI Coverage and RoI Quality Index) designed specifically for medical report generation assessment. Models trained on VietPET-RoI demonstrated substantial improvements, achieving 19.7\% and 4.7\% gains  
over the strongest baseline method (M3D-LaMed) in BLEU-4 and ROUGE-L, alongside 15.5\% and 45.8\% improvements in clinical metrics, indicating enhanced reliability and reduced hallucination.

\vspace{-0.2cm}
\section*{Limitations}

Despite its fine-grained RoI annotations, VietPET-RoI has several limitations. First, the dataset scale remains modest with 200 patients from a single institution, which may affect generalization and introduce potential demographic bias. Expanding to multiple medical centers would enhance diversity. Second, the RoI annotation process requires specialized medical expertise and is time-intensive (averaging 15-30 minutes per case), limiting rapid scaling. Third, due to the dataset's uniqueness, we lack established benchmarks for comprehensive evaluation, particularly for tasks such as automatic RoI segmentation. While we have established baseline methods for report generation, developing benchmarks for intermediate tasks remains an important research direction. Finally, the 3D architecture of HiRRA demands substantial computational resources, which may limit deployment in resource-constrained settings. We believe VietPET-RoI establishes an important foundation for region-aware PET/CT research and encourages the community to develop additional methods and benchmarks in this domain.
\section*{Ethical Considerations}

This study was conducted in strict adherence to medical research ethics standards and the ACM Code of Ethics. The protocol was reviewed and approved by the Ethics Committee of the data-providing institution.

\textbf{Informed Consent and Privacy}: Due to the retrospective nature of the study, the Institutional Review Board (IRB) granted a formally validated waiver of informed consent, determining that the research poses minimal risk to subjects. All patient data underwent a rigorous de-identification process validated by institutional authorities to ensure complete anonymity. Personal Protected Information (PPI) was removed in compliance with international standards (e.g., HIPAA Safe Harbor). Consequently, there are no privacy risks associated with the public release of this dataset.

\textbf{Intended Use and Fairness}: VietPET-RoI is released exclusively for non-commercial research to support medical AI development. It is not a diagnostic tool and must not replace clinical judgment; any deployment requires further rigorous clinical validation. We acknowledge that data sourced from a single institution may contain inherent demographic biases. To mitigate potential harm, we advise future researchers to validate models on diverse external populations before clinical consideration.

The dataset will be publicly released for non-commercial research purposes under specified usage terms upon paper acceptance. 

Given the complete anonymization process, there are no privacy concerns associated with public release. All ethical considerations have been reviewed and validated by institutional authorities, ensuring compliance with medical research ethics standards.

\section*{Acknowledgements}
This research is partially supported by Toray Industries (H.K.) Ltd. Vietnam. We also extend our deepest gratitude to the physicians and staff at the collaborating hospital for their expertise in data curation and RoI annotation. This research was made possible by the contributions of patients whose anonymized medical records form the basis of this dataset.

\bibliography{anthology}

@article{gatidis2022whole,
  title={A whole-body FDG-PET/CT dataset with manually annotated tumor lesions},
  author={Gatidis, Sergios and Hepp, Tobias and Fr{\"u}h, Marcel and La Foug{\`e}re, Christian and Nikolaou, Konstantin and Pfannenberg, Christina and Sch{\"o}lkopf, Bernhard and K{\"u}stner, Thomas and Cyran, Clemens and Rubin, Daniel},
  journal={Scientific Data},
  volume={9},
  number={1},
  pages={601},
  year={2022},
  publisher={Nature Publishing Group UK London}
}

@article{li2023llava,
  title={Llava-med: Training a large language-and-vision assistant for biomedicine in one day},
  author={Li, Chunyuan and Wong, Cliff and Zhang, Sheng and Usuyama, Naoto and Liu, Haotian and Yang, Jianwei and Naumann, Tristan and Poon, Hoifung and Gao, Jianfeng},
  journal={Advances in Neural Information Processing Systems},
  volume={36},
  pages={28541--28564},
  year={2023}
}

@article{beyer2004acquisition,
  title={Acquisition protocol considerations for combined PET/CT imaging},
  author={Beyer, Thomas and Antoch, Gerald and M{\"u}ller, Stefan and Egelhof, Thomas and Freudenberg, Lutz S and Debatin, J{\"o}rg and Bockisch, Andreas},
  journal={Journal of Nuclear Medicine},
  volume={45},
  number={1 suppl},
  pages={25S--35S},
  year={2004},
  publisher={Society of Nuclear Medicine}
}

@inproceedings{pieper20043d,
  title={3D Slicer},
  author={Pieper, Steve and Halle, Michael and Kikinis, Ron},
  booktitle={2004 2nd IEEE international symposium on biomedical imaging: nano to macro (IEEE Cat No. 04EX821)},
  pages={632--635},
  year={2004},
  organization={IEEE}
}

@article{dhouib2021roi,
  title={ROI-based compression strategy of 3D MRI brain datasets for wireless communications},
  author={Dhouib, D and Na{\"\i}t-Ali, A and Olivier, C and Naceur, MS},
  journal={IRBM},
  volume={42},
  number={3},
  pages={146--153},
  year={2021},
  publisher={Elsevier}
}

@article{ortiz2014automatic,
  title={Automatic ROI selection in structural brain MRI using SOM 3D projection},
  author={Ortiz, Andr{\'e}s and G{\'o}rriz, Juan M and Ram{\'\i}rez, Javier and Martinez-Murcia, Francisco J and Alzheimer's Disease Neuroimaging Initiative},
  journal={PloS one},
  volume={9},
  number={4},
  pages={e93851},
  year={2014},
  publisher={Public Library of Science San Francisco, USA}
}

@inproceedings{chen2024internvl,
  title={Internvl: Scaling up vision foundation models and aligning for generic visual-linguistic tasks},
  author={Chen, Zhe and Wu, Jiannan and Wang, Wenhai and Su, Weijie and Chen, Guo and Xing, Sen and Zhong, Muyan and Zhang, Qinglong and Zhu, Xizhou and Lu, Lewei and others},
  booktitle={Proceedings of the IEEE/CVF conference on computer vision and pattern recognition},
  pages={24185--24198},
  year={2024}
}

@article{zhu2025internvl3,
  title={Internvl3: Exploring advanced training and test-time recipes for open-source multimodal models},
  author={Zhu, Jinguo and Wang, Weiyun and Chen, Zhe and Liu, Zhaoyang and Ye, Shenglong and Gu, Lixin and Tian, Hao and Duan, Yuchen and Su, Weijie and Shao, Jie and others},
  journal={arXiv preprint arXiv:2504.10479},
  year={2025}
}

@inproceedings{shi2025medm,
  title={Medm-vl: What makes a good medical lvlm?},
  author={Shi, Yiming and Yang, Shaoshuai and Zhu, Xun and Wang, Haoyu and Fu, Xiangling and Li, Miao and Wu, Ji},
  booktitle={International Workshop on Agentic AI for Medicine},
  pages={290--299},
  year={2025},
  organization={Springer}
}

@article{xin2025med3dvlm,
  title={Med3dvlm: An efficient vision-language model for 3d medical image analysis},
  author={Xin, Yu and Ates, Gorkem Can and Gong, Kuang and Shao, Wei},
  journal={IEEE Journal of Biomedical and Health Informatics},
  year={2025},
  publisher={IEEE}
}

@article{bai2024m3d,
  title={M3d: Advancing 3d medical image analysis with multi-modal large language models},
  author={Bai, Fan and Du, Yuxin and Huang, Tiejun and Meng, Max Q-H and Zhao, Bo},
  journal={arXiv preprint arXiv:2404.00578},
  year={2024}
}

@article{nguyen2026toward,
  title={Toward a vision-language foundation model for medical data: Multimodal dataset and benchmarks for vietnamese pet/ct report generation},
  author={Nguyen, Tien and Nguyen, Dac and Nguyen, Trung Thanh and Thao Nguyen, Truong and Pham, Hieu and Barthelemy, Johan and Quan, Tran Minh and Nguyen, Quoc Viet Hung and Nguyen, Thanh Tam and Son, Mai and others},
  journal={Advances in Neural Information Processing Systems},
  volume={38},
  year={2026}
}

@inproceedings{zhang2026pet2rep,
  title={PET2Rep: Towards vision-language model-drived automated radiology report generation for positron emission tomography},
  author={Zhang, Yichi and Zhang, Wenbo and Ling, Zehui and Feng, Gang and Peng, Sisi and Chen, Deshu and Liu, Yuchen and Zhang, Hongwei and Wang, Shuqi and Li, Lanlan and others},
  booktitle={Proceedings of the AAAI Conference on Artificial Intelligence},
  volume={40},
  pages={12897--12906},
  year={2026}
}

@article{jiao2025vision,
  title={Vision-Language Models for Automated 3D PET/CT Report Generation},
  author={Jiao, Wenpei and Shang, Kun and Li, Hui and Yan, Ke and Zhang, Jiajin and Yang, Guangjie and Guo, Lijuan and Wan, Yan and Yang, Xing and Jin, Dakai and Xie, Zhaoheng},
  journal={arXiv preprint arXiv:2511.20145},
  year={2025}
}

@misc{lin2017featurepyramidnetworksobject,
      title={Feature Pyramid Networks for Object Detection}, 
      author={Tsung-Yi Lin and Piotr Dollár and Ross Girshick and Kaiming He and Bharath Hariharan and Serge Belongie},
      year={2017},
      eprint={1612.03144},
      archivePrefix={arXiv},
      primaryClass={cs.CV},
      url={https://arxiv.org/abs/1612.03144}, 
}

@inproceedings{li2023blip,
  title={Blip-2: Bootstrapping language-image pre-training with frozen image encoders and large language models},
  author={Li, Junnan and Li, Dongxu and Savarese, Silvio and Hoi, Steven},
  booktitle={International conference on machine learning},
  pages={19730--19742},
  year={2023},
  organization={PMLR}
}

@article{hu2022lora,
  title={Lora: Low-rank adaptation of large language models.},
  author={Hu, Edward J and Shen, Yelong and Wallis, Phillip and Allen-Zhu, Zeyuan and Li, Yuanzhi and Wang, Shean and Wang, Liang and Chen, Weizhu and others},
  journal={Iclr},
  volume={1},
  number={2},
  pages={3},
  year={2022}
}

@misc{brody2022attentivegraphattentionnetworks,
      title={How Attentive are Graph Attention Networks?}, 
      author={Shaked Brody and Uri Alon and Eran Yahav},
      year={2022},
      eprint={2105.14491},
      archivePrefix={arXiv},
      primaryClass={cs.LG},
      url={https://arxiv.org/abs/2105.14491}, 
}

@article{wu2025towards,
  title={Towards generalist foundation model for radiology by leveraging web-scale 2d\&3d medical data},
  author={Wu, Chaoyi and Zhang, Xiaoman and Zhang, Ya and Hui, Hui and Wang, Yanfeng and Xie, Weidi},
  journal={Nature Communications},
  volume={16},
  number={1},
  pages={7866},
  year={2025},
  publisher={Nature Publishing Group UK London}
}

@article{townsend2008petct,
  title={Multimodality imaging of structure and function},
  author={Townsend, David W},
  journal={Physics in Medicine \& Biology},
  volume={53},
  number={4},
  pages={R1},
  year={2008},
  publisher={IOP Publishing}
}

@article{kinahan2006petct,
  title={X-ray-based attenuation correction for positron emission tomography/computed tomography scanners},
  author={Kinahan, Paul E and Hasegawa, Bruce H and Beyer, Thomas},
  journal={Seminars in Nuclear Medicine},
  volume={33},
  number={3},
  pages={166--179},
  year={2003},
  publisher={Elsevier},
  note={Key kept as requested, original paper 2003}
}

@article{von2009petct,
  title={Integrated PET/CT: current applications and future directions},
  author={Von Schulthess, Gustav K and Steinert, Hans C and Hany, Thomas F},
  journal={Radiology},
  volume={251},
  number={3},
  pages={708--736},
  year={2009},
  publisher={Radiological Society of North America}
}

@article{muzi2015data,
  title={Data from rider lung pet-ct},
  author={Muzi, Peter and Wanner, Michelle and Kinahan, Paul},
  journal={(No Title)},
  year={2015},
  publisher={The Cancer Imaging Archive}
}

@misc{Goel_LangExtract_2025,
author = {Goel, Akshay},
doi = {10.5281/zenodo.17015089},
license = {Apache-2.0},
month = nov,
title = {{LangExtract}},
url = {https://github.com/google/langextract},
version = {1.1.1},
year = {2025}
}

@article{kuhn1955hungarian,
  title={The Hungarian method for the assignment problem},
  author={Kuhn, Harold W},
  journal={Naval research logistics quarterly},
  volume={2},
  number={1-2},
  pages={83--97},
  year={1955},
  publisher={Wiley Online Library}
}

@article{ju2024graph,
  title={Graph neural network model for prediction of non-small cell lung cancer lymph node metastasis using protein--protein interaction network and 18F-FDG PET/CT radiomics},
  author={Ju, Hyemin and Kim, Kangsan and Kim, Byung Il and Woo, Sang-Keun},
  journal={International Journal of Molecular Sciences},
  volume={25},
  number={2},
  pages={698},
  year={2024},
  publisher={MDPI}
}

@article{kazmierski2021lymph,
  title={Lymph node graph neural networks for cancer metastasis prediction},
  author={Kazmierski, Michal and Haibe-Kains, Benjamin},
  journal={arXiv preprint arXiv:2106.01711},
  year={2021}
}

@article{gu2026medvh,
  title={MedVH: Toward systematic evaluation of hallucination for large vision language models in the medical context},
  author={Gu, Zishan and Chen, Jiayuan and Liu, Fenglin and Yin, Changchang and Zhang, Ping},
  journal={Advanced Intelligent Systems},
  volume={8},
  number={1},
  pages={2500255},
  year={2026},
  publisher={Wiley Online Library}
}

@article{messina2022survey,
  title={A survey on deep learning and explainability for automatic report generation from medical images},
  author={Messina, Pablo and Pino, Pablo and Parra, Denis and Soto, Alvaro and Besa, Cecilia and Uribe, Sergio and And{\'\i}a, Marcelo and Tejos, Cristian and Prieto, Claudia and Capurro, Daniel},
  journal={ACM Computing Surveys (CSUR)},
  volume={54},
  number={10s},
  pages={1--40},
  year={2022},
  publisher={ACM New York, NY}
}

@article{waite2019analysis,
  title={Analysis of perceptual expertise in radiology--current knowledge and a new perspective},
  author={Waite, Stephen and Grigorian, Arkadij and Alexander, Robert G and Macknik, Stephen L and Carrasco, Marisa and Heeger, David J and Martinez-Conde, Susana},
  journal={Frontiers in human neuroscience},
  volume={13},
  pages={213},
  year={2019},
  publisher={Frontiers Media SA}
}

@inproceedings{xie2025medtrinity,
  title={Medtrinity-25m: A large-scale multimodal dataset with multigranular annotations for medicine},
  author={Xie, Yunfei and Zhou, Ce and Gao, Lang and Wu, Juncheng and Li, Xianhang and Zhou, Hong-Yu and Liu, Sheng and Xing, Lei and Zou, James Y and Xie, Cihang and others},
  booktitle={International Conference on Learning Representations},
  volume={2025},
  pages={6036--6060},
  year={2025}
}

@inproceedings{boecking2022making,
  title={Making the most of text semantics to improve biomedical vision--language processing},
  author={Boecking, Benedikt and Usuyama, Naoto and Bannur, Shruthi and Castro, Daniel C and Schwaighofer, Anton and Hyland, Stephanie and Wetscherek, Maria and Naumann, Tristan and Nori, Aditya and Alvarez-Valle, Javier and others},
  booktitle={European conference on computer vision},
  pages={1--21},
  year={2022},
  organization={Springer}
}

@article{de2025padchest,
  title={Padchest-gr: A bilingual chest X-ray dataset for grounded radiology report generation},
  author={de Castro, Daniel Coelho and Bustos, Aurelia and Bannur, Shruthi and Hyland, Stephanie L and Bouzid, Kenza and Wetscherek, Maria Teodora and S{\'a}nchez-Valverde, Maria Dolores and Jaques-P{\'e}rez, Lara and P{\'e}rez-Rodr{\'\i}guez, Lourdes and Takeda, Kenji and others},
  journal={NEJM AI},
  volume={2},
  number={7},
  pages={AIdbp2401120},
  year={2025},
  publisher={Massachusetts Medical Society}
}

@article{zhang2024hierarchical,
  title={Hierarchical medical image report adversarial generation with hybrid discriminator},
  author={Zhang, Junsan and Cheng, Ming and Cheng, Qiaoqiao and Shen, Xiuxuan and Wan, Yao and Zhu, Jie and Liu, Mengxuan},
  journal={Artificial Intelligence in Medicine},
  volume={151},
  pages={102846},
  year={2024},
  publisher={Elsevier}
}

@misc{hamamci2024ct2repautomatedradiologyreport,
      title={CT2Rep: Automated Radiology Report Generation for 3D Medical Imaging}, 
      author={Ibrahim Ethem Hamamci and Sezgin Er and Bjoern Menze},
      year={2024},
      eprint={2403.06801},
      archivePrefix={arXiv},
      primaryClass={eess.IV},
      url={https://arxiv.org/abs/2403.06801}, 
}

@misc{maqbool2025petarlocalizedfindingsgeneration,
      title={PETAR: Localized Findings Generation with Mask-Aware Vision-Language Modeling for PET Automated Reporting}, 
      author={Danyal Maqbool and Changhee Lee and Zachary Huemann and Samuel D. Church and Matthew E. Larson and Scott B. Perlman and Tomas A. Romero and Joshua D. Warner and Meghan Lubner and Xin Tie and Jameson Merkow and Junjie Hu and Steve Y. Cho and Tyler J. Bradshaw},
      year={2025},
      eprint={2510.27680},
      archivePrefix={arXiv},
      primaryClass={cs.CV},
      url={https://arxiv.org/abs/2510.27680}, 
}

@misc{sellergren2026medgemmatechnicalreport,
      title={MedGemma Technical Report}, 
      author={Andrew Sellergren and Sahar Kazemzadeh and Tiam Jaroensri and Atilla Kiraly and Madeleine Traverse and Timo Kohlberger and Shawn Xu and Fayaz Jamil and Cían Hughes and Charles Lau and Justin Chen and Fereshteh Mahvar and Liron Yatziv and Tiffany Chen and Bram Sterling and Stefanie Anna Baby and Susanna Maria Baby and Jeremy Lai and Samuel Schmidgall and Lu Yang and Kejia Chen and Per Bjornsson and Shashir Reddy and Ryan Brush and Kenneth Philbrick and Mercy Asiedu and Ines Mezerreg and Howard Hu and Howard Yang and Richa Tiwari and Sunny Jansen and Preeti Singh and Yun Liu and Shekoofeh Azizi and Aishwarya Kamath and Johan Ferret and Shreya Pathak and Nino Vieillard and Ramona Merhej and Sarah Perrin and Tatiana Matejovicova and Alexandre Ramé and Morgane Riviere and Louis Rouillard and Thomas Mesnard and Geoffrey Cideron and Jean-bastien Grill and Sabela Ramos and Edouard Yvinec and Michelle Casbon and Elena Buchatskaya and Jean-Baptiste Alayrac and Dmitry Lepikhin and Vlad Feinberg and Sebastian Borgeaud and Alek Andreev and Cassidy Hardin and Robert Dadashi and Léonard Hussenot and Armand Joulin and Olivier Bachem and Yossi Matias and Katherine Chou and Avinatan Hassidim and Kavi Goel and Clement Farabet and Joelle Barral and Tris Warkentin and Jonathon Shlens and David Fleet and Victor Cotruta and Omar Sanseviero and Gus Martins and Phoebe Kirk and Anand Rao and Shravya Shetty and David F. Steiner and Can Kirmizibayrak and Rory Pilgrim and Daniel Golden and Lin Yang},
      year={2026},
      eprint={2507.05201},
      archivePrefix={arXiv},
      primaryClass={cs.AI},
      url={https://arxiv.org/abs/2507.05201}, 
}

\clearpage

\appendix

\begin{center}
    \Large \textbf{Supplementary Material}
\end{center}
\vspace{1em}
\vspace{-0.5cm}
\section{Detailed Dataset Construction Process}
\label{sec:appendix_dataset}
\subsection{Dataset Collection and Pre-processing}
The dataset was curated by oncology specialists at a leading public hospital in Vietnam to ensure clinical accuracy and authenticity. After screening, 200 cancer cases spanning the six most common malignancies were selected. These cases were drawn from examinations performed between 2017 and 2019 and cover a diverse range of sex, age, and body habitus, supporting the development of models with improved generalization (see Table~\ref{tab:dataset_summary}).

Clinical PET/CT at the site is acquired using several protocols \cite{beyer2004acquisition} that differ in image format and slice count. To reduce protocol-related variability, all studies were standardized to a single whole-body protocol with 313 axial slices from the head to the upper thighs. The raw PET/CT volumes were exported in DICOM format, containing rich metadata such as patient information, body weight, standardized uptake values (SUV), and other physiological attributes. The corresponding reports were collected as physician-typed Word documents (DOC), following a semi-structured template with fields including “Gender,” “Examination Date,” “Indication,” “Clinical History,” “Impression,” and “Image Description.”.

After collection, a series of preprocessing steps was applied to obtain model-ready inputs. For the textual reports, all patient-identifiable information was removed to satisfy ethical and privacy requirements. The de-identified reports were then converted from DOCX to a standardized JSON schema, enabling direct alignment with the corresponding images during training. Automatic extraction was followed by manual review to correct spelling or formatting inconsistencies and to ensure the reliability of key fields.
\begin{table}[th!]
    \caption{Schema of the region-of-interest (RoI) annotation template.}
    \label{tab:RoI_template}
    \rowcolors{2}{gray!8}{white} 
    \begin{tabular}{p{0.27\linewidth} p{0.65\linewidth}}
        \toprule
        \rowcolor{gray!20}
        \textbf{Field} & \textbf{Description} \\
        \midrule
        Anatomic region & Anatomical location of the finding (e.g., left supraclavicular, liver segment~VI, right ovary). \\
        Lesion type & Type of lesion or physiological structure; physiological findings are explicitly labeled as ``physiological''. \\
        Size & In-plane lesion size in millimeters. \\
        SUVmax & Maximum standardized uptake value of the lesion, when available. \\
        Density & Qualitative CT density / attenuation (e.g., hypodense, soft-tissue density). \\
        Morphology & Morphological appearance (e.g., rounded, irregular, thickened wall, nodule). \\
        FDG uptake & Qualitative FDG metabolism (e.g., increased or non-increased uptake). \\
        Top-3 possible diseases & Three most likely diagnoses, listed in descending order of likelihood. \\
        Top-3 further examinations & Three recommended follow-up tests or diagnostic procedures, ordered by priority. \\
        Physical region & Coarse body region (1 = head--neck, 2 = chest, 3 = abdomen--pelvis). \\
        Note & Optional free-text comments if needed. \\
        \bottomrule
    \end{tabular}
\end{table}
For the 3D PET/CT data, the DICOM volumes were converted to NumPy array (NPY) format. To increase the number of samples and refine the supervision signal, each whole-body study (image plus report) was further divided into three anatomically meaningful regions: head–neck, chest, and abdomen–pelvis. Neighboring regions were defined with a 15-slice overlap to preserve continuity and avoid missing important findings at segment boundaries. Concretely, the head–neck region corresponds to the first 25\% of slices in the scan; the chest region starts 25\% below the last slice of the head–neck segment; and the abdomen–pelvis region covers the remaining slices from the chest down to the pelvis. These proportions were validated by clinical experts and empirically verified to ensure that each regional volume fully covers all relevant RoIs for that anatomical area. This regional segmentation increases the effective sample size and improves the alignment between visual content and textual descriptions, enabling more precise region-level learning and, in turn, better overall model performance.

\subsection{Region-of-Interest (RoI) Annotation}

From the three-region split of each of the 200 cases, we obtained 600 image--report pairs for region-level supervision. We then collaborated with eight physicians to annotate regions of interest (RoIs) and provide structured descriptions. Annotation was performed in 3DSlicer \cite{pieper20043d}: every abnormal finding or clinically relevant physiological structure mentioned in the report was localized on the corresponding PET/CT volume and marked with a 3D bounding box. For each RoI, annotators completed a compact structured template that records key clinical attributes, as summarized in Table~\ref{tab:RoI_template}.
3DSlicer \cite{pieper20043d} exports the coordinates of each RoI together with its structured string description in this format, resulting in fine-grained, region-level supervision for every study. By linking spatial localization with rich clinical semantics, the annotations provide strong guidance for report generation models to focus on clinically important regions and help reduce irrelevant or hallucinated content in the generated reports \cite{dhouib2021RoI, ortiz2014automatic}.

\section{Detailed RoI Metrics}
\label{sec:detail_RoI_metric}

\subsection{Ground Truth RoI Annotation Schema}
To capture the semantic richness of PET/CT reports, our dataset utilizes a structured annotation schema for Regions of Interest (RoIs). Each ground truth RoI is represented as a structured string containing 11 distinct clinical attributes. The annotation format is defined as follows:

\vspace{0.5em}
\begin{mdframed}[backgroundcolor=gray!8, linewidth=0.5pt, linecolor=gray!40, roundcorner=2pt, innertopmargin=6pt, innerbottommargin=6pt, innerleftmargin=6pt, innerrightmargin=6pt]
\small \texttt{[Anatomic Region] - [Lesion Type] - [Size] - [SUVmax] - [Density] - [Morphology] - [FDG Uptake] - [Top-3 Differential Diagnoses] - [Top-3 Recommended Examinations] - [Physical Region ID] - [Clinical Note]}
\end{mdframed}

\vspace{0.5em}
\noindent \textbf{Example Instance:}
\vspace{0.5em}
\begin{mdframed}[backgroundcolor=blue!5, linewidth=0.5pt, linecolor=blue!20, roundcorner=2pt, innertopmargin=6pt, innerbottommargin=6pt, innerleftmargin=6pt, innerrightmargin=6pt]
\small \texttt{[Cecum] - [Focal hypermetabolism] - [Unclear] - [12.3] - [Soft tissue density] - [Focal] - [Very intense hypermetabolism] - [Colon cancer (cecum), Inflammatory bowel disease (Crohn's disease), Appendicitis/Abscess] - [Colonoscopy and biopsy, Abdominal MRI/CT, Blood tests] - [3] - [Very intense focal FDG uptake (SUVmax 12.3) in the cecum. Highly suggestive of colon cancer...]}
\end{mdframed}

\begin{figure*}[ht]
    \centering
    \includegraphics[width=\textwidth]{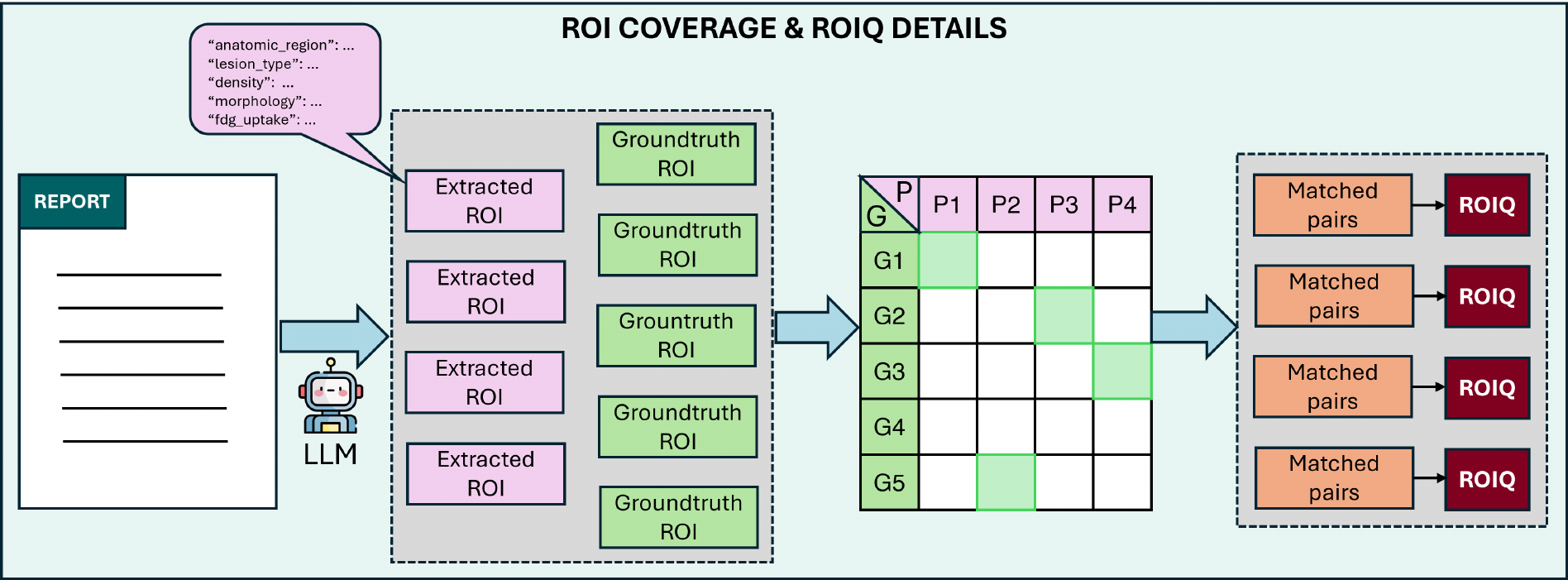} 
    \caption{\textbf{Overview of our proposed clinical evaluation protocol.} We utilize an LLM-based framework to extract structured clinical attributes from reports. RoI Coverage is quantified by aligning predicted and ground-truth RoIs via embedding-based Hungarian matching. For aligned pairs, the RoI Quality Index (RoIQ) measures semantic fidelity, strictly enforcing anatomical and lesion-type correctness.}
    \label{fig:appendix_RoI}
    \vspace{-0.4cm}
\end{figure*}

\subsection{Information Extraction Framework}
To evaluate generation quality, we employ \textit{LangExtract}~\cite{Goel_LangExtract_2025}, an LLM-based extraction framework designed to parse unstructured generated reports into structured RoI objects. Specifically, we utilize Gemini-2.5-Pro as the backbone LLM. For every predicted sentence in a report, the framework extracts the information into the following structured JSON format:
\vspace{0.5em}
\begin{mdframed}[backgroundcolor=gray!5, linewidth=0.5pt, linecolor=gray!40, roundcorner=2pt, innertopmargin=6pt, innerbottommargin=6pt, innerleftmargin=6pt, innerrightmargin=6pt]
\small \texttt{\{}\\
\hspace*{1em}\texttt{"extraction\_text": "...", \textcolor{gray}{// The original sentence text}}\\
\hspace*{1em}\texttt{"anatomic\_region": "...",}\\
\hspace*{1em}\texttt{"lesion\_type":     "...",}\\
\hspace*{1em}\texttt{"density":         "...",}\\
\hspace*{1em}\texttt{"morphology":      "...",}\\
\hspace*{1em}\texttt{"fdg\_uptake":     "..."}\\
\texttt{\}}
\end{mdframed}

\vspace{0.5em}
 
While the ground-truth RoI schema contains 10 clinical attributes and 1 data management identifier, the primary objective of our clinical metrics (RoI Coverage and RoIQ) is to assess the correctness and semantic fidelity of the localized lesions. After consultation with clinical partners, we determined that only the five extracted fields above directly reflect this capability, as they characterize the observable properties of a lesion. The remaining six attributes were excluded for the following methodological reasons:

\begin{itemize}
    \setlength\itemsep{0em}
    \item \textbf{Data Management Artifacts:} The \textit{physical region code} is a preprocessing identifier used solely to split full-body scans into anatomical segments during dataset construction, rather than a clinically observable feature. 
    \item \textbf{Unstructured Context:} \textit{Additional notes} consist of free-text commentary derived from the complete clinical report. These contain auxiliary observations extending beyond the scope of individual RoIs and cannot be systematically evaluated through field-level matching.
    \item \textbf{Downstream Diagnostic Reasoning:} \textit{Top-3 possible diseases} and \textit{top-3 further examinations} are synthesized from the entire report context. They exhibit high inter-annotator variability and are downstream from RoI identification-they depend on correct lesion characterization but do not define it.
    \item \textbf{Numerical Regression Constraints:} \textit{Size} and \textit{SUVmax} are quantitative measurements requiring precise numerical computation from raw NumPy imaging arrays. Evaluating these fields would conflate the model's spatial identification ability with numerical regression, which current VLMs cannot reliably perform from visual tokens alone. 
\end{itemize}

Therefore, these five attributes isolate and directly measure the core capability our metrics are designed to assess: whether the model correctly identifies and semantically characterizes lesions as observable entities.

\subsection{Quantitative Evaluation: RoI Coverage}
We introduce \textbf{RoI Coverage} to evaluate lesion localization as a set-based detection problem. We compare the set of Ground Truth RoIs ($G$) against the set of Predicted RoIs ($P$) extracted by LangExtract.

\subsubsection{Embedding and Similarity Calculation}
For each RoI, we isolate five comparable text fields: \textit{anatomic region, lesion type, density, morphology,} and \textit{fdg uptake}. We utilize a clinical embedding model to convert the text value of each field into a vector representation. For a predicted RoI $p_i \in P$ and a ground truth RoI $g_j \in G$, we calculate the similarity score $S(p_i, g_j)$ based on the cosine similarities of their constituent fields (excluding \texttt{extraction\_text}).

\subsubsection{Optimal Matching (Hungarian Algorithm)}
We construct a similarity matrix $M \in \mathbb{R}^{|P| \times |G|}$ where each entry $M_{i,j}$ represents the similarity between $p_i$ and $g_j$. To resolve the alignment, we apply the Hungarian Algorithm~\cite{kuhn1955hungarian} (Linear Sum Assignment) to find the optimal set of pairs that maximizes the total similarity score. A pair $(p_i, g_j)$ is considered a valid match (True Positive) only if their similarity score exceeds a predefined threshold $\tau$:
\begin{equation}\nonumber
    \text{Match}(p_i, g_j) = \mathds{1}[M_{i,j} \geq \tau]
\end{equation}

\subsubsection{Classification Metrics}
Based on valid matches, we quantify \textbf{RoI Coverage} using True Positives ($TP$), False Positives ($FP$), and False Negatives ($FN$):
\begin{itemize}
    \item $TP$: Successfully matched ground truth RoIs.
    \item $FP = |P| - TP$: Predicted RoIs that did not match any ground truth (hallucinations).
    \item $FN = |G| - TP$: Ground truth RoIs that were not detected.
\end{itemize}

Standard classification metrics are then computed as:
\begin{equation}\nonumber
\begin{gathered}
    \text{Precision} = \frac{TP}{TP + FP}, \quad
    \text{Recall} = \frac{TP}{TP + FN} \\
    F1 = 2 \cdot \frac{\text{Precision} \cdot \text{Recall}}{\text{Precision} + \text{Recall}}
\end{gathered}
\label{eq:metrics}
\end{equation}

\subsection{Qualitative Evaluation: RoI Quality Index (RoIQ)}
For every successfully matched pair of RoIs, we introduce the \textbf{RoI Quality Index (RoIQ)} to assess the semantic accuracy of the generated details. This metric prioritizes the correct identification of critical attributes (region and lesion type) before evaluating descriptive attributes (density, morphology, FDG uptake).

Let $S_{\text{region}}, S_{\text{lesion}}, S_{\text{density}}, S_{\text{morphology}}, S_{\text{uptake}}$ denote the similarity scores for the five fields. We define the set of secondary attributes as $\mathcal{A} = \{S_{\text{density}}, S_{\text{morphology}}, S_{\text{uptake}}\}$. Since descriptive attributes may be absent in the ground truth, we compute the mean score only over the subset of non-empty attributes, $\mathcal{A}_{\text{valid}} \subseteq \mathcal{A}$. The RoIQ is defined as:

\begin{equation}\nonumber
    \text{RoIQ} = \sqrt{S_{\text{region}} \cdot S_{\text{lesion}}} \times \left( \frac{1}{|\mathcal{A}_{\text{valid}}|} \sum_{s \in \mathcal{A}_{\text{valid}}} s \right)
\end{equation}

\noindent Note: If a descriptive attribute is missing in the ground truth, it is excluded from $|\mathcal{A}_{\text{valid}}|$ to prevent penalizing the model for not generating non-existent features.

\subsection{Sensitivity Analysis of Similarity Threshold ($\tau$)}
\label{sec:appendix_tau_sensitivity}
We conducted an experiment to evaluate the sensitivity of the similarity threshold $\tau$ across representative values on a held-out development set of 60 samples (240 RoI pairs) via grid search over $\tau \in [0.5, 0.95]$ with a step size of 0.05.

According to the experimental results presented in Table~\ref{tab:tau_sensitivity}, decreasing $\tau$ to 0.55 increases Recall (0.502) but substantially degrades Match Quality ($RoIQ = 0.501$), indicating that the metric becomes overly permissive and accepts semantically distinct pairs as valid matches. Conversely, increasing $\tau$ to 0.80 causes a significant drop in both Recall (0.337) and F1 (0.295), penalizing clinically valid semantic variations such as synonymous radiological expressions. 

The selected $\tau = 0.70$ achieves the best trade-off between comprehensive lesion coverage and semantic precision, and is therefore adopted as our default threshold.

\begin{table}[ht]
\centering
\small
\caption{Sensitivity analysis of the similarity threshold ($\tau$) on the validation set (60 samples).}
\label{tab:tau_sensitivity}
\begin{tabular}{lcccc}
\toprule
\textbf{Threshold ($\tau$)} & \textbf{F1} & \textbf{Recall} & \textbf{Precision} & \textbf{RoIQ} \\
\midrule
0.55 & 0.435 & 0.502 & 0.368 & 0.501 \\
\textbf{0.70 (Selected)} & \textbf{0.371} & \textbf{0.422} & \textbf{0.332} & \textbf{0.573} \\
0.80 & 0.295 & 0.337 & 0.264 & 0.525 \\
\bottomrule
\end{tabular}
\end{table}
\section{Implementation Details}
\label{app:settings}

We implemented our framework using PyTorch and the HuggingFace \texttt{transformers} library. All experiments were conducted on a single NVIDIA A100 (80GB) GPU. We provide the detailed training hyperparameters and optimization settings in Table~\ref{tab:hyperparams}.

\subsection{Progressive Training Strategy}

Our model employs a four-stage progressive training curriculum to ensure stable convergence and prevent catastrophic forgetting. Starting from vision encoder pretraining on large-scale medical data, we sequentially align global visual features, integrate region-specific information, and perform end-to-end finetuning. This staged approach allows each component to specialize in its designated task before joint optimization, significantly improving training stability compared to end-to-end training from scratch.

In the final end-to-end finetuning stage (Stage 4), all model components are jointly optimized with carefully tuned learning rates to balance exploration and exploitation. We adopt Low-Rank Adaptation (LoRA)~\cite{hu2022lora} for parameter-efficient finetuning of the large language model (Qwen3-8B), applying rank-32 adapters with $\alpha=64$ and dropout rate of 0.1 to the query, key, value, and output projection layers. This reduces trainable parameters by 99.7\% while maintaining model expressiveness. To prevent overfitting, we employ early stopping with a patience of 5 epochs, monitoring validation loss with a minimum improvement threshold of 0.001.

\subsection{Training Hyperparameters}

\begin{table}[h]
    \centering
    \small
    \caption{Training hyperparameters for end-to-end finetuning (Stage 4). Earlier stages use similar configurations with component-specific learning rates detailed in the supplementary materials.}
    \begin{tabular}{l|c}
        \toprule
        \textbf{Hyperparameter} & \textbf{Value} \\
        \midrule
        \multicolumn{2}{c}{\textit{Optimization}} \\
        \midrule
        Optimizer & AdamW\\
        Learning Rate (LoRA) & $5 \times 10^{-6}$ \\
        Learning Rate (Vision) & $1 \times 10^{-6}$ \\
        LR Scheduler & Cosine Annealing \\
        Warmup Ratio & 10\% of total steps \\
        Weight Decay & $0.02$ \\
        Adam $\beta_1$, $\beta_2$ & $(0.9, 0.999)$ \\
        Adam $\epsilon$ & $1 \times 10^{-8}$ \\
        \midrule
        \multicolumn{2}{c}{\textit{Regularization}} \\
        \midrule
        Max Gradient Norm & $1.0$ \\
        Dropout (LoRA) & $0.1$ \\ 
        \midrule
        \multicolumn{2}{c}{\textit{Batch Configuration}} \\
        \midrule
        Batch Size per GPU & 1 \\
        Effective Batch Size & 2 \\
        Total Epochs & 10 \\
        \midrule
        \multicolumn{2}{c}{\textit{Hardware \& Efficiency}} \\
        \midrule
        Mixed Precision & BF16 \\
        GPU & 1 $\times$ NVIDIA A100 (80GB) \\
        Number of Workers & 8 \\
        \bottomrule
    \end{tabular}
    \label{tab:hyperparams}
\end{table}

\noindent\textbf{Learning Rate Strategy.} We employ a conservative learning rate strategy to prevent catastrophic forgetting of pretrained knowledge. Vision components (encoder, feature extractors) use an order-of-magnitude lower learning rate ($1 \times 10^{-6}$) compared to LoRA adapters ($5 \times 10^{-6}$), allowing the language model to adapt to multimodal inputs while preserving visual representations learned in earlier stages. A cosine annealing scheduler with 10\% linear warmup steps gradually reduces learning rates, facilitating smooth convergence.

\noindent\textbf{Regularization.} To mitigate overfitting on the relatively small medical dataset, we apply multiple regularization techniques: (1) weight decay of 0.02 on all parameters except biases and layer normalization weights; (2) gradient clipping with maximum norm of 1.0 to stabilize training; and (3) dropout of 0.1 in LoRA adapters. These techniques collectively prevent the model from memorizing training samples while maintaining generalization capability.

\subsection{Loss Function}

Our model is optimized using the standard autoregressive language modeling objective with cross-entropy loss:
\begin{equation}
    \mathcal{L}_{\text{LM}} = -\frac{1}{T}\sum_{t=1}^{T} \log P(y_t \mid y_{<t}, \mathbf{x})
\end{equation}
where $\mathbf{x} = [\mathbf{x}_{\text{demo}}, \mathbf{x}_{\text{global}}, \mathbf{x}_{\text{RoI}}]$ represents the multimodal input comprising patient demographics, global visual context ($M=32$ query vectors), and RoI-specific features (graph-enhanced visual embeddings), $y_t$ denotes the target token at position $t$, and $T$ is the target sequence length. The loss is computed only on target tokens, with prompt tokens masked by setting their labels to -100. During training, we apply teacher forcing where ground-truth tokens are used as input for predicting subsequent tokens, enabling efficient parallel computation across the sequence dimension.

\end{document}